\documentclass[orivec]{llncs}

\usepackage[utf8]{inputenc}
\usepackage{amsmath, amssymb}
\usepackage{graphicx}
\usepackage{xspace}
\usepackage{tikz,subfigure}
\usetikzlibrary{arrows,shapes,shapes.multipart,snakes,automata,backgrounds,petri,positioning,shadows,matrix,decorations.pathmorphing, decorations.pathreplacing, decorations.markings, fit,positioning,calc,backgrounds,shapes.misc,arrows.meta,fit}
\usepackage{todonotes}
\usepackage[defblank]{paralist}
\usepackage{booktabs}
\usepackage[capitalise]{cleveref}
\usepackage{url}

\usepackage{algorithm}
\usepackage{algorithmic}

\usepackage{xspace}
\newcommand{\mathname}[1]{\ensuremath{\textit{#1}}}

\makeatletter
\g@addto@macro\normalsize{%
\setlength{\abovecaptionskip}{0pt}
\setlength{\belowcaptionskip}{-10pt}
\setlength\abovedisplayskip{3pt}
\setlength\belowdisplayskip{3pt}
\setlength\abovedisplayshortskip{3pt}
\setlength\belowdisplayshortskip{3pt}
}
\makeatother

\DeclareMathOperator*{\argmin}{argmin}

\newcommand{\A}{\ensuremath{\mathcal{A}}}

\newcommand{\C}{\ensuremath{\mathcal{C}}}

\newcommand{\E}{\ensuremath{\mathcal{E}}}

\newcommand{\G}{\ensuremath{\mathcal{G}}}
\renewcommand{\H}{\ensuremath{\mathcal{H}}}

\renewcommand{\L}{\ensuremath{\mathcal{L}}}

\newcommand{\N}{\ensuremath{\mathcal{N}}}

\renewcommand{\P}{\ensuremath{\mathcal{P}}}

\newcommand{\W}{\ensuremath{\mathcal{W}}}

\newcommand{\true}{\top}
\newcommand{\set}[1]{\{#1\}}                      % set

\newcommand{\tup}[1]{\langle #1\rangle}            % tuple

\newcommand{\card}[1]{|{#1}|}                     % cardinality of a set

\newcommand{\reals}{\mathbb{R}\xspace}

\newcommand{\ltr}[1]{\ensuremath{\overset{#1}{\longrightarrow}}\xspace}
\newcommand{\lra}{\ensuremath{\longrightarrow}\xspace}

\newcommand{\trace}{\sigma}

\newcommand{\const}[1]{\ensuremath{\mathsf{#1}}\xspace}
\newcommand{\EV}{\ensuremath{\mathcal{V_\E}}\xspace}

\newcommand{\limp}{\rightarrow}

\newcommand{\Var}{\mathname{Var}}

\newcommand{\net}{D}

\newcommand{\guard}[1]{[\![#1]\!]}
\newcommand{\pre}[1]{{{}^\bullet{#1}}} % input places of transition
\newcommand{\post}[1]{{{#1}^\bullet}} % output places of transition
\newcommand{\goto}[1]{\ensuremath[{#1}\rangle}

\newcommand{\mpdeclare}{\ensuremath{\textsc{MP-Declare}}\xspace}
\newcommand{\lmpdeclare}{\ensuremath{\textsc{LMP-Declare}}\xspace}

\newcommand{\ra}{\ensuremath{\rightarrow}}

\newcommand{\ltlf}{\ensuremath{\textsc{LTL}_f}\xspace}

\newcommand{\lnext}{\ensuremath{\mathbf{X}}}

\newcommand{\luntil}{\ensuremath{\mathbf{U}}}

\newcommand{\lglobally}{\ensuremath{\mathbf{G}}}
\newcommand{\lfuture}{\ensuremath{\mathbf{F}}}

\newcommand{\LTLf}{\ensuremath{\text{LTL}_f}\xspace}

\newcommand{\declare}{Declare\xspace}

\newcommand{\activity}[1]{\texttt{#1}}

\newcommand{\gfa}{GFA\xspace}

\newcommand{\cfunc}{\ensuremath{cost}\xspace}
\newcommand{\cbest}{\ensuremath{\cfunc_{best}}\xspace}
\newcommand{\ccur}{\ensuremath{\cfunc_{cur}}\xspace}

\definecolor{deepblue}{HTML}{0C3B80}
\definecolor{deepgreen}{HTML}{2EA601}
\definecolor{lightOrange}{HTML}{FFA03C}
\definecolor{darkOrange}{HTML}{F1800A}
\definecolor{lightBlue}{HTML}{0174CD}
\definecolor{greenF}{HTML}{2CBB5C}
\definecolor{cyan}{HTML}{86A6D5}
\definecolor{darkred}{HTML}{8B0000}

\def\DTZU {2ex}

\tikzstyle{taskfg} = [
  text=deepblue,
]

\tikzstyle{taskbg} = [
  fill=cyan!50,
]

\tikzstyle{taskline} = [
  draw=deepblue,
]

\tikzstyle{taskstyle} = [
  ultra thick,
  taskfg,
  taskbg,
  taskline
]

\tikzstyle{task} = [
  rectangle,
  minimum width=12mm,
  minimum height=10mm,
  taskstyle
]

\tikzstyle{smalltask} = [
  rectangle,
  minimum width=8mm,
  minimum height=6mm,
  taskstyle,
  very thick
]

\tikzstyle{constraint} = [
  ultra thick,
  taskline,
  taskbg
]

\tikzstyle{response} = [
  constraint,
  *-triangle 60
]

\tikzstyle{precedence} = [
  constraint,
  *triangle 60-
]
\tikzstyle{succession} = [
  constraint,
  *triangle 60-*
]

\tikzstyle{respondedexistence} = [
  constraint,
  *-
]

\tikzstyle{coexistence} = [
  constraint,
  *-*
]

\tikzstyle{negationconstraint} = [
  constraint,
  postaction={decorate,decoration={markings,
   mark=at position .5 with {\arrow[xshift=0.15*\DTZU]{Bar[width=1.5*\DTZU]}},
   mark=at position .5 with {\arrow[xshift=-0.15*\DTZU]{Bar[width=1.5*\DTZU]}}
  }}
]

\tikzstyle{notcoexistence} = [
  negationconstraint,
  coexistence,
]

\tikzstyle{negationresponse} = [
  constraint,
  response,
  postaction={decorate,decoration={markings,
   mark=at position .5 with {\arrow[xshift=0.15*\DTZU]{Bar[width=1.5*\DTZU]}},
   mark=at position .5 with {\arrow[xshift=-0.15*\DTZU]{Bar[width=1.5*\DTZU]}}
  }}
]

\tikzstyle{negationsuccession} = [
  constraint,
  succession,
  postaction={decorate,decoration={markings,
   mark=at position .65 with {\arrow[xshift=0.15*\DTZU]{Bar[width=1.5*\DTZU]}},
   mark=at position .65 with {\arrow[xshift=-0.15*\DTZU]{Bar[width=1.5*\DTZU]}}
  }}
]

\tikzstyle{exclusivechoice} = [
  constraint,
  -,
  postaction={decorate,decoration={markings,
   mark=at position .5 with {\arrow[xshift=0.5*\DTZU]{Diamond[width=1*\DTZU]}} 
  }}
]

\tikzstyle{choice} = [
  constraint,
  -,
  postaction={decorate,decoration={markings,
   mark=at position .5 with {\arrow[xshift=0.5*\DTZU]{Diamond[width=1*\DTZU]}},
    mark=at position .5 with {\arrow[xshift=0.28*\DTZU,white,scale=.6]{Diamond[width=1*\DTZU]}} 
  }}
]

\tikzstyle{circ} = [
  solid,
  text=black
]

\tikzstyle{timeline} = [
  -,
  thin,
  ]
  
\tikzstyle{snapshot} = [
  -,
  thin,
  densely dotted,
  ]

\tikzstyle{objnode} = [
  inner sep=2pt,
  circle,
  ultra thick,
]

\tikzstyle{tobj} = [
  objnode,
  taskfg,
  taskbg,
  taskline
]

\tikzstyle{cobj} = [
  objnode,
  classfg,
  classbg,
  classline
]

\tikzstyle{link} = [
  solid,
  -angle 60
]

\newcommand{\plname}[1]{\mathtt{#1}}
\newcommand{\trname}[1]{\mathtt{#1}}
\tikzstyle{place}=[circle,thick,draw=black,fill=white,minimum size=7mm,font=\fontsize{9}{144}\selectfont]
\tikzstyle{transition}=[rectangle,thick,draw=black,fill=gray!20,minimum size=7mm]

\title{Monitoring Hybrid Process Specifications with Conflict Management: The Automata-theoretic Approach}
\author{ Anti Alman$^1$ \and Fabrizio Maria Maggi$^2$ \and Marco Montali$^2$ \and Fabio Patrizi$^3$ \and Andrey Rivkin$^2$}
\authorrunning{Alman, Maggi, Montali, Patrizi, Rivkin}
\institute{%
$^1$University of Tartu, Tartu, Estonia\\Email: anti.alman@ut.ee\\
$^2$Free University of Bozen-Bolzano, Bolzano, Italy\\Email: \{maggi,montali,andrey\}@inf.unibz.it\\
$^3$Sapienza University of Rome, Rome, Italy\\Email: patrizi@diag.uniroma1.it
}

\begin{document}

\maketitle

\begin{abstract}
Business process monitoring approaches have thus far mainly focused on monitoring the execution of a process with respect to a single process model. However, in some cases it is necessary to consider multiple process specifications simultaneously. In addition, these specifications can be procedural, declarative, or a combination of both. For example, in the medical domain, a clinical guideline describing the treatment of a specific disease cannot account for all possible co-factors that can coexist for a specific patient and therefore additional constraints may need to be considered. In some cases, these constraints may be incompatible with clinical guidelines, therefore requiring the violation of either the guidelines or the constraints. In this paper, we propose a solution for monitoring the interplay of hybrid process specifications expressed as a combination of (data-aware) Petri nets and temporal logic rules. During the process execution, if these specifications are in conflict with each other, it is possible to violate some of them. The monitoring system is equipped with a violation cost model according to which the system can recommend the next course of actions in a way that would either avoid possible violations or minimize the total cost of violations.
\end{abstract}

%!TEX root = ../main.tex

\section{Introduction}
\label{sec:introduction}
A key functionality of any process-aware information system is {\it monitoring}~\cite{DBLP:journals/is/LyMMRA15}. Monitoring concerns the ability to verify at runtime whether an ongoing process execution conforms to the corresponding process model. This runtime form of conformance checking allows to detect, and therefore handle, deviations appearing in ongoing process instances. However, in several scenarios, different process specifications must be valid during the process execution and the monitoring system should take into consideration all of them and their interplay.

One such scenario would be the treatment of a patient having co-morbid conditions. In this case, the standard treatment procedures for each condition can be specified using procedural models, while additional knowledge, such as harmful drug interactions, can be specified using declarative constraints \cite{DBLP:conf/bpm/BottrighiCMMMT11}.
%The constraints may limit the treatment options or require performing additional activities.
Note that, the interplay of process specifications can generate conflicts during the process execution \cite{DBLP:conf/bpm/BottrighiCMMMT11,DBLP:conf/biostec/PiovesanTD18}. For example, by making a decision that, based on a procedural model, will lead to administering a drug that the patient is allergic to. To be able to take informed decisions in these situations, experts responsible for the execution of such process(es) need to be promptly alerted about the presence of conflicts.

In this paper, we present a monitoring approach with respect to multiple process specifications, each of which may also include conditions on the data perspective. In particular, we use data Petri nets (DPNs)~\cite{DBLP:journals/computing/MannhardtLRA16} for procedural models and Linear Temporal Logic over finite traces (\LTLf)~\cite{DeVa13} for declarative models (additionally supporting the \LTLf based modeling language \mpdeclare~\cite{DBLP:conf/edoc/PesicSA07,DBLP:journals/eswa/BurattinMS16}). This allows us to capture sophisticated forms of scoping and interaction among the different process specifications (that is, the different elicited DPNs and declarative constraints), going beyond what is captured so far in the literature, and providing a full logic-based characterization of the so-resulting hybrid processes \cite{DBLP:conf/bpm/BottrighiCMMMT11,DBLP:conf/biostec/PiovesanTD18,DBLP:journals/artmed/TerenzianiMT01}. These models differ from loosely coupled hybrid models \cite{DBLP:journals/is/AndaloussiBSKW20}, in that the different process specifications all interact with each other at the same level of abstraction.

Of particular importance, in our monitoring approach, is the early detection of conflicts among process specifications, arising when the process is in a state where at least one specification will eventually be violated. This aspect has been considered before with a purely declarative approach \cite{MMWV11,MWMV11,DDGM14} but never applied to a hybrid setting.

In our context, there are three main novel challenges that need to be addressed. First and foremost, we need to tackle the infinity induced by the presence of data, which in general leads to undecidability of monitoring. In our specific setting, we show that we can recast data abstraction techniques studied for verification of DPNs \cite{LeoniFM18,LeoniFM20} so as to produce finitely representable monitors based on finite-state automata. Second, we need to homogeneously construct monitors for constraints and DPNs, and define how to combine them into a unique, global monitor for conflict detection; we do so by recasting the standard notion of automata product, producing a global monitor that conceptually captures a hybrid model where DPNs and constraints are all simultaneously applied (i.e. all DPNs are executed concurrently, while checking the validity of constraints). Third, we need to handle situations where the global monitor returns a permanent violation (due to the explicit violation of a process specification, or the presence of a conflict), but distinguishing among different continuations is still relevant as they may lead to violate different process specifications. Assuming a violation cost is given for each specification, we show how to augment our monitors with the ability of returning the best-possible next events, that is, those keeping the overall violation cost at the minimum possible.

The remainder of this paper is structured as follows. \Cref{sec:scenario} provides an example monitoring scenario. \Cref{sec:components} and \Cref{sec:monitoring} introduce the necessary preliminaries and the monitoring approach respectively. \Cref{sec:conclusion} concludes the paper.

%!TEX root = ../main.tex

\section{Example Scenario}
\label{sec:scenario}

Consider the following real-life scenario, where a patient with co-morbidities is simultaneously treated with different guidelines: a guideline for peptic ulcer (PU) and a guideline for venous thromboembolism (VT). More specifically, we are considering two tiny, yet relevant fragments of the guideline models presented in~\cite{DBLP:conf/biostec/PiovesanTD18}. The two fragments are represented in \figurename~\ref{fig:net-ex} using DPNs (recalled in Section~\ref{sec:components}).
%(\figurename~\ref{fig:CGs}).

%(\figurename~\ref{fig:CGs}).

%\begin{figure*}[!t]
%    \centering
%	\includegraphics[width=.7\textwidth]{figures/CGs}
%	\caption{Clinical guidelines for peptic ulcer (PU) and for venous thromboembolism (VT).}
%    \label{fig:CGs}
%\end{figure*}
%\todo[inline]{Change Figure~\ref{fig:CGs}: improve resolution or redesign}

When PU starts, the helicobacter pylori test is executed. Based on the test result, different therapies are chosen: amoxicillin administration in case of positive test, gastric acidity reduction otherwise. Afterwards, the peptic ulcer is evaluated to estimate the effects of the therapy.

VT requires an immediate intervention, chosen among three different possibilities based on the situation of the specific patient at hand. Mechanical intervention uses devices that prevent the proximal propagation or embolization of the thrombus into the pulmonary circulation, or involves the removal of the thrombus. The other two possibilities are an anticoagulant therapy based on warfarin, or a thrombolytic therapy.

The interaction between amoxicillin therapy (in the PU procedure) and warfarin therapy (in the VT procedure) is usually avoided in medical practice, since amoxicillin increases the anticoagulant effect of warfarin, raising the risk of bleedings. Therefore, in cases where the PU and VT procedures are performed simultaneously, the medical practice suggests specifying that \textsl{amoxicillin therapy} and \textsl{warfarin therapy} cannot coexist (declarative constraint C). Such a constraint $C$ is an example of background medical knowledge rule \cite{DBLP:conf/bpm/BottrighiCMMMT11}.

Based on these specifications, if helicobacter pylori is tested positive and anticoaugulant is chosen to deal with venous thromboembolism, then there is a conflict between C and the two guidelines. In this outlier, but possible situation there would be three alternatives:
\begin{enumerate}
    \item Violating PU (by skipping the amoxicillin therapy);
  \item Violating VT (by using an alternative anticoagulant);
  \item Violating C (giving priority to the two guidelines).
\end{enumerate}

Informing medical experts about the presence of a conflict is crucial to help them in assessing the current situation, ponder the implications of one choice over the others, and finally make an informed decision. One can go beyond mere information, by also presenting the violation severity of the different alternatives. This can be done by assigning violation costs to the process specifications (in our case, PU, VT, and C). In this case, we can assume that skipping the amoxicillin therapy is rather costly, given the lack of viable alternatives for treating peptic ulcer in case of helicobacter pylori. Instead, violating the VT procedure has a lower cost, given the existence of other anticoagulants (e.g., heparin) that may be less effective but do not interact strongly with amoxicillin. Additionally, constraint C comes with the highest violation cost as complications such as serious bleeding should definitely be avoided.

%We can also express durative actions by distinguishing the start event and the complete event of an activity. In this case, to better specify the interaction between \textsl{amoxicillin therapy} and \textsl{warfarin therapy}, we split them into \textsl{amoxicillin therapy-start} and \textsl{amoxicillin therapy-complete} and \textsl{warfarin therapy-start} and \textsl{warfarin therapy-complete}. This requires redefining constraint C to specify that \textsl{amoxicillin therapy-start} and \textsl{warfarin therapy-start} can never coexist in the same case. \todo{This works only if we have data! And we need more than the data we have here\ldots!}

In our approach, we can also deal with more sophisticated (meta-)constraints \cite{DBLP:conf/bpm/GiacomoMGMM14} that impose conditions on the process execution depending on the truth value of other constraints. Within our example scenario, we can for example specify a meta-constraint dictacting that if constraint C gets violated, then at least we expect that \textsl{warfarin therapy} is executed \emph{after} \textsl{amoxicillin therapy}, to reduce the risk of a harmful interaction of warfarin and amoxicillin.

\section{Process Components}
\label{sec:components}
In this section, we define the models used to specify declarative and procedural data-aware process components, by relying on Multi Perspective-\declare (MP-\declare) \cite{DBLP:journals/eswa/BurattinMS16,DDGM14,MaMB19}) and data Petri nets (DPNs \cite{DBLP:journals/computing/MannhardtLRA16,LeoniFM18}). 

\subsection{Events and Conditions}

We start by fixing some preliminary notions related to events and traces. 
An \emph{event signature} is a tuple $\tup{n,A}$, where:
$n$ is the \emph{activity} name and $A=\set{a_1,\ldots,a_\ell}$ 
is the set of event \emph{attribute (names)}.
We assume a finite set $\E$ of event signatures, each having a distinct name (thus we can simply refer to an event signature $\tup{n,A}$ using its name $n$).
By $\N_\E=\bigcup_{\tup{n,A}\in\E} n$ we denote the set of all event names from $\E$
and by $\A_\E=\bigcup_{\tup{n,A}\in\E} A$ the set of all attribute 
names occurring in $\E$.

Each event comes with a name matching one of the names in $\N_\E$, and provides actual values for the attributes of the corresponding signature. In the context of this paper, attributes range over reals equipped with comparison predicates (simpler types such as strings with equality and booleans can be seamlessly encoded).
 
\begin{definition}[Event]
\label{def:event}
An \emph{event} of event signature $\tup{n,A}$ is a 
pair $e=\tup{n,\nu}$ where $\nu:A\mapsto\reals$ is a 
total function assigning a real value to each attribute in $A$.
\end{definition}
%%Given an event instance $e=\tup{n,\nu}$, we denote
%%by $e.n$ the name of the class of $e$, by $e.A$ the set $n.A$, and by $e.a$
%%the value assigned by $\nu$ to attribute $a\in e.A$.
%%Finally, we use $\EV$ to denote the set of all possible events of some
%%class in $\E$.
As usual, sequences of events form (process) traces.
\begin{definition}
\label{def:trace}
  A \emph{trace} over a set $\E$ of event signatures is a finite sequence 
  $\trace=e_1\cdots e_\ell$, where each $e_i$ is an event of some signature in $\E$.
\end{definition}
By $\card{\trace}$ we denote  the \emph{length} of $\trace$. For 
$1\leq i\leq \card{\trace}$, we define $\trace(i)\doteq e_i$.

%%This is, in essence, the standard semantics of propositional formulae,
%%where each event defines a propositional interpretation.
%%Obviously, an event can satisfy different formulas. 
%\todo[inline]{\textbf{Definition (just in case).} An \emph{event} is a couple $(b,\alpha)$, where $b\in A$ is an activity label and $\alpha$ is a (possibly partial) variable assignment. 
%We use $\E$ to denote a set of events. Given $\E$, we define a \emph{(log) trace} $\logtrace\in\E^*$ as a sequence of events from $\E$. An \emph{event log} $L\in\M(\E^*)$ is a multiset of log traces, where  $\M(\E^*)$ denotes the set of multisets over $\E^*$.}

%!TEX root = ../../main.tex

\subsection{Multi-Perspective \declare with Local Conditions}

To represent declarative process components, we resort to a multi-perspective variant of the well-known \declare language~\cite{DBLP:conf/edoc/PesicSA07}. A \declare model describes \emph{constraints} that must be satisfied throughout the process execution. Constraints, in turn, are based on \emph{templates}. Templates are patterns that define parameterized classes of properties, and constraints are their concrete instantiations.
The template semantics is formalized using Linear Temporal Logic over finite traces (LTL$_f$) \cite{MPVC10}. 
 
%A central shortcoming of languages like \declare is the fact that templates are not directly capable of expressing the connection between the behavior and other perspectives of the process. In particular, when events carry a data payload, it is important to specify not only how the event names temporally relate to each other, but also which conditions should/may hold over the data they carry. Several \emph{multi-perspective} variants of \declare have been defined for this purpose, differing in the types of data and corresponding conditions they support \cite{DDGM14,DBLP:journals/eswa/BurattinMS16,MaMB19}.
%%

In this work, we consider temporal constraints enriched with boolean combinations of attribute-to-constant comparisons. The resulting language closely resembles that of variable-to-constant conditions in \cite{LeoniFM18}, thus providing a good basis for combining declarative constraints with procedural models expressed with DPNs. %\cite{LeoniFM18}.

\begin{definition}
  \label{def:condition}
A condition $\varphi$ over a set  $\E$ of event signatures is an expression of the form:
\[\varphi := x\mid a\odot c\mid\lnot\varphi\mid\varphi\land\varphi,\]
where:
	\begin{inparaenum}[]
		\item $x\in\N_\E$;
		\item $a\in\A_\E$;
		\item ${\odot}\in\set{<,=,>}$;
		\item $c\in\reals$.
	\end{inparaenum}
\end{definition}
%$\L_\E$ is a propositional language over propositions of the form $s$ and $a\odot c$. 
Conditions of the form $a\odot c$ and $x$ are called \emph{atomic}.
We define the usual abbreviations:
$\varphi_1\lor\varphi_2\doteq \lnot(\lnot \varphi_1\land\lnot\varphi_2)$;
$\varphi_1\ra\varphi_2\doteq \lnot\varphi_1\lor\varphi_2$;
$a\leq c\doteq\lnot(a>c)$; $a\geq c\doteq\lnot (a<c)$; and 
$a\neq c\doteq\lnot (a=c)$.
In addition, we denote by $\L_\E$ the language of conditions over $\E$.

Conditions of $\L_\E$ are interpreted over events as follows.
\begin{definition}
\label{def:cond-semantics}
We inductively define when a condition $\varphi$ is \emph{satisfied} by an event 
$e=\tup{n,\nu}$, written $e \models \varphi$, as follows:
	\begin{itemize}
		\item $e\models x$ iff $x=n$;
		\item $e\models a\odot c$ iff $\nu(a)$ is defined and $\nu(a)\odot c$;
		\item $e\models\lnot\varphi$ iff $e\not\models\varphi$;
		\item $e\models\varphi_1\land\varphi_2$ iff $e\models\varphi_1$ and
		$e\models\varphi_2$.
	\end{itemize}
\end{definition}

We are now ready to define LMP-\declare constraints, that is, \mpdeclare constraints with local conditions. Their syntactic and semantic definition basically corresponds to that of \ltlf formulae with conditions as atomic formulae, interpreted over traces of the form given in Definition~\ref{def:trace}.

\begin{definition}
\label{def:lmp-declare}
  An \emph{LMP-\declare constraint} is an expression of the form:
\[
\Phi := \true\mid \varphi \mid \lnext\,\Phi \mid \Phi_1 \luntil\,\Phi_2 \mid \neg \Phi \mid \Phi_1 \land \Phi_2 
\] 
where $\varphi$ is a condition from $\L_\E$ (cf.~Definition~\ref{def:condition}).
\end{definition}
As in standard \ltlf, $\lnext$ denotes the \emph{strong next} operator (which requires the existence of a next state where the inner formula holds), while $\luntil$ stands for \emph{strong until} (which requires the right-hand formula to eventually hold, forcing the left-hand formula to hold in all intermediate states).

\begin{definition}
  \label{def:lmp-semantics}
We inductively define when an LMP-\declare constraint $\Phi$ is \emph{satisfied} 
by a trace $\trace$ at position $1\leq i\leq \card{\trace}$, written $\trace,i \models \Phi$, as follows:
\begin{compactitem}
	\item $\trace,i \models \true$;
	\item $\trace,i \models \varphi$ iff $\trace(i) \models \varphi$ according to Definition~\ref{def:cond-semantics};
	\item $\trace,i \models \Phi_1 \land \Phi_2$ iff $\trace,i \models \Phi_1$ and $\trace,i \models \Phi_2$;
	\item $\trace,i \models \neg \Phi$ iff $\trace,i \not\models \Phi_1$;
	\item $\trace,i \models \lnext\,\Phi$ iff $i < \card{\trace}$ and $\trace,i+1 \models \Phi$;
    \item $\trace,i \models \Phi_1 \luntil \Phi_2$ iff there exists $j$, $1\leq j\leq \card{\trace}$, 
    s.t.~$\trace,j\models\Phi_2$ and for every $k$, $1\leq k\leq j-1$, we have $\trace,k\models \Phi_1$. 
\end{compactitem}
\end{definition}
We define the usual abbreviations:
$\Phi_1 \lor \Phi_2 \doteq \neg (\neg \Phi_1 \land \neg \Phi_2)$;
$\Phi_1 \limp \Phi_2 \doteq \neg \Phi_1 \lor \Phi_2$;
$\lfuture\,\Phi = true \luntil\,\Phi$ (\emph{eventually}); 
and $\lglobally\,\Phi = \neg \lfuture\,\neg \Phi$ (\emph{globally}).

\begin{example}
Consider two event signatures $\tup{\activity{a},\set{x,y}}$ 
and $\tup{\activity{b},\set{z}}$. The \emph{negation response} LMP-\declare constraint
\[
\lglobally
(
	\activity{a} \limp \neg \lnext\lfuture
	(\activity{b} \land z > 10)
)
\]	
captures that whenever event $\activity{a}$ occurs then $\activity{b}$ cannot later occur with its attribute $z$ carrying a value greater than $10$.
\end{example}

%Techniques based on finite-state automata are then used to carry out several reasoning tasks over the resulting constraints, such as monitoring \cite{MMWV11}. More recently, Declare patterns have been extended with so-called \emph{metaconstraints}, that is, temporal constraints that predicate about the state of other constraints, where a state indicates that the target constraint is permanently violated, temporarily violated, permanently satisfied, or temporarily satisfied. Notably, metaconstraints can be formalized in Linear Dynamic Logic over finite traces (LDL$_f$), which extends LTL$_f$ but retains the same automata-based techniques for monitoring~\cite{DBLP:conf/bpm/GiacomoMGMM14}.

%!TEX root = ../../main.tex

\subsection{Data Petri nets}
\label{sec:dpn}

%%DPNs extend traditional place-transition nets with the possibility to manipulate scalar case variables, which are used as basic
%%building blocks to constrain the evolution of the process through data-aware constraints (called \emph{guards}) assigned to
%%transitions.
We define data Petri nets (DPNs) by adjusting~ \cite{DBLP:journals/computing/MannhardtLRA16,LeoniFM18} to our needs. In particular, our definition needs to accommodate the fact that a monitored trace will be matched against multiple process components (which will be the focus of Section~\ref{sec:monitoring}).

Let $\E$ be a finite set of event signatures. The language $\G_\E$ of \emph{guards} $\gamma$
over $\E$ is defined as follows:
\[\gamma := a\odot c\mid\lnot\gamma\mid\gamma_1\land\gamma_2.\]
Observe that $\G_\E$ is the sub-language of conditions over $\E$, i.e., $\L_\E$, with
formulas $\gamma$ not mentioning event names.
We can then specialize the notion of satisfaction to guards, by considering
only the assignment to the event attributes. Namely, given an assignment
$\alpha: \A_\E\to \reals$ and an atomic condition $a\odot c$, we have that
$\alpha\models a\odot c$ iff $\alpha(a)\odot c$. Boolean combinations of atomic conditions are defined as usual. Given a condition $\gamma$, we denote by $\Var(\gamma)$ the set of attributes mentioned therein.

%%and having the
%%negation symbol only in front of atomic guards, i.e., formulas of the form $a\odot c$.
%%We denote the set of all atomic guards over $\E$ as $\AG_\E$.
%%
%%Given a set of variables $V$,
%%an \emph{atomic guard (over $V$)} is an expression of the form $v\odot k$, s.t.
%%$v\in V$, $k\in \mathbb{R}$ and $\odot\in\set{<,>,=,\neq,\geq,\leq}$.
%%
%%A \emph{guard} $\gamma$ is a boolean combination of atomic guards from $\AG$, with $\G$
%%denoting the set of all such guards.
%%
%%

%%Since DPN guards inspect and update case variables, we associate a given set $V$ of variables with two corresponding disjoint
%%sets of annotated variables $V^r = \{v^r \mid v\in V\}$ and $V^w = \{v^w \mid v\in V\}$. Without loss of generality, we consider that
%%constraints and guards are defined on set $V^r\cup V^w$.

\begin{definition}%[Petri Net with Data (DPN)]
A \emph{Petri net with data and variable-to-constant conditions} (DPN) over a set $\E$ of event signatures is a tuple
$\net = \tup{P, T, F, l, V, r, w}$, where:
\begin{compactitem}
	\item $P$ and $T$ are two finite disjoint sets of \emph{places} and \emph{transitions}, respectively;
	\item $F:(P \times T)\cup(T \times P)\to\mathbb{N}$ is the net's \emph{flow relation};
	\item $l:T\rightarrow \N_\E\cup\set{\tau}$ is a total \emph{labeling} function assigning
	a label from $\N_\E\cup\set{\tau}$ to every transition $t\in T$, with $\tau$ denoting a \emph{silent} transition.
	\item $V\subseteq \A_\E$ is the set of net's \emph{variables};
	\item $r: T \to \G_\E$ and $w: T \to \G_\E$ are two total \emph{read} and \emph{write}
	\emph{guard-assignment} functions, mapping every transition $t\in T$ into
	a read and write guard from $\G_\E$.
%	s.t.~$\Var(\gamma_t^r),\Var(\gamma_t^w)\subseteq V$ and
%	$\Var(\gamma_t^r)\cap\Var(\gamma_t^w)=\emptyset$, i.e., the write and read guards
%	have no common variables.
\end{compactitem}
\end{definition}
 We respectively call $\Var_r(t)$ and $\Var_w(t)$ the sets of
	$t$'s \emph{read} and \emph{write} variables, as a shortcut notation for $\Var(r(t))$ and $\Var(w(t))$. Given a place or a transition $x\in P \cup T$ of $D$, the \emph{preset} and the
\emph{postset} of $x$ are, respectively, the sets
$\pre{x}=\set{y\mid F(y,x)>0}$ and
$\post{x}:=\set{y\mid F(x,y)>0}$.

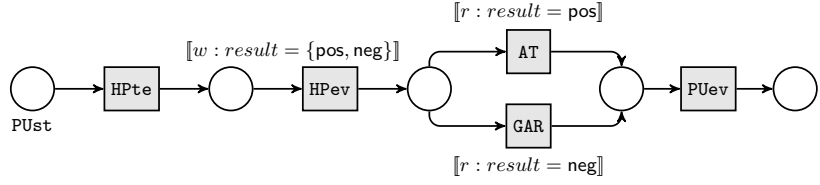
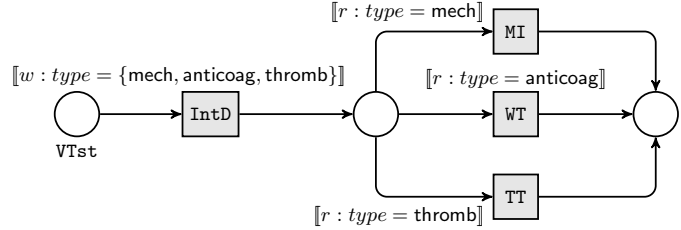
\begin{figure*}[t!]
		\centering
        \subfigure
        [DPN for peptic ulcer treatment]
        {
        \label{fig:net-ex1}
				\resizebox{.9\textwidth}{!}{
				\begin{tikzpicture}[->,>=stealth',auto,x=25mm,y=1.0cm,node distance=16mm 	and 16mm,thick]
			
				\node[place,label=below:$\plname{PUst}$,xshift=2mm] (start) at (0,0) {};
				\node[transition,right of = start] (hpte) {$\trname{HPte}$};
				\node[place,right of = hpte] (p1) {};
				
				\node[transition,right of = p1] (hpev) {$\trname{HPev}$};
				\node[above of = hpev, xshift=-6mm, yshift=-14mm,label=$\guard{w:
				result={\{\const{pos},\const{neg}\}}}$]  (g1)  {};

				\node[place,right of = hpev] (p2) {};
				\node[transition,right of = p2,yshift=6mm,label=above:$\guard{r:result=\const{pos}}$] (am) {$\trname{AT}$};

				\node[transition,right of = p2,yshift=-6mm,label=below:$\guard{r:result =\const{neg}}$] (gar) {$\trname{GAR}$};

				\node[place,right of = p2,xshift=1.5cm] (p3) {};
				\node[transition,right of = p3,xshift=-2mm] (eval) {$\trname{PUev}$};
				\node[place,right of = eval,xshift=-2mm] (end) {};

				\path[]
   				 (start) edge (hpte)
   				 (hpte) edge (p1)
   				 (p1) edge (hpev)
   				 (hpev) edge (p2);
   				
   				 \draw[rounded corners=5pt]
   				 (p2) |- (am);
   				
   				 \draw[rounded corners=5pt]
   				 (p2) |- (gar);
   				
   				 \draw[rounded corners=5pt]
   				 (gar) -| (p3);
   				
   				 \draw[rounded corners=5pt]
   				 (am) -| (p3);  				
   				
   				 \path[]
   				 	(p3) edge (eval)
   				 (eval) edge (end)
    			;

  			\end{tikzpicture}
 }
}
     \hfill
     \subfigure[DPN for thromboembolism treatment]
     {
     \label{fig:net-ex2}
     			\resizebox{.75\textwidth}{!}{
     			\begin{tikzpicture}[->,>=stealth',auto,x=25mm,y=1.0cm,node distance=16mm 	and 16mm,thick]

				\node[place,right=1.5cm of end,label=below:$\plname{VTst}$] (start) {};
				\node[transition,right of = start, xshift=5mm] (tr) {$\trname{IntD}$};
				\node[above of = tr, xshift=-5mm,yshift=-14mm,label=$\guard{w:type=\const{\{mech,anticoag,thromb\}}}$]  (g1)  {};

				\node[place,right of = tr, xshift=10mm] (p1) {};

				\node[transition,right of = p1,xshift=6mm,yshift=1.3cm] (mi) {$\trname{MI}$};
				\node[left = 0mm of mi, anchor=east, yshift=3mm]  (g2)  {$\guard{r:type=\const{mech}}$};
				\node[transition,right of = p1,xshift=6mm,yshift=0cm] (wt) {$\trname{WT}$};
				\node[above of = wt, yshift=-14.5mm, label=$\guard{r:type=\const{anticoag}}$]  (g3)  {};

				\node[transition,right of = p1,xshift=6mm,yshift=-1.3cm] (tt) {$\trname{TT}$};
				\node[left= 0mm of tt, anchor=east, yshift=-3mm]  (g4)  {$\guard{r:type=\const{thromb}}$};

				\node[place,right of = wt, xshift=6mm] (end) {};

\draw
(start) edge (tr);

\draw
(tr) edge (p1);

\draw[rounded corners=5pt]
(p1) |- (mi);

\draw
(p1) edge (wt);

\draw[rounded corners=5pt]
(p1) |- (tt);

\draw[rounded corners=5pt]
(mi) -| (end);

\draw
(wt) edge (end);

\draw[rounded corners=5pt]
(tt) -| (end);
  			\end{tikzpicture}
  			}}
  	\caption{DPN representations for the peptic ulcer (left) and venous thromboembolism (right) clinical guideline fragments. We use prefixes \emph{r:} and \emph{w:} to distinguish read and write guards respectively. Trivial, true guards are omitted for brevity.}
  	\label{fig:net-ex}
\end{figure*}

\begin{example}
\label{ex:dpn}
Figure~\ref{fig:net-ex} shows two DPNs encoding the two clinical guideline fragments discussed in Section~\ref{sec:scenario}. The two figures employ string constants, which can be easily encoded into dedicated real numbers to fit our formal definition.
\end{example}

We turn to the DPN execution semantics. A \emph{state} of a DPN $\net = (P, T, F, l, V, r, w)$ over $\E$ is a pair $(M,\alpha)$, where:
\begin{compactitem}
	\item $M:P\to\mathbb{N}$ is a total \emph{marking} function, assigning a number $M(p)$ of \emph{tokens}
		to every place $p\in P$;
	\item $\alpha:V\to\reals$ is a total \emph{variable valuation} (function)
		assigning a real value to every variable in $V$.
\end{compactitem}
Every state, together with a (variable) valuation $\beta$ inducing an update over (some of) the net variables,
yields a set of \emph{enabled} transitions,
which can be \emph{fired} to progress the net. This requires to augment the usual notions of enablement/firing by considering also the read and write guards.
%After a transition fires a new state
%is reached, with a new  corresponding marking and valuation.

%%
\begin{definition}
\label{def:dpn-transition-firing}
	Consider a DPN $\net = (P, T, F, l, V, r, w)$.
	Transition $t \in T$ is \emph{enabled}
	in state $(M,\alpha)$ under partial valuation $\beta:V\nrightarrow\reals$, denoted $(M,\alpha)[t,\beta\rangle$, iff:
	\begin{compactitem}
		\item $\beta$ is defined on all variables $v\in \Var_r(t)\cup \Var_w(t)$;
		\item for every $v\in \Var_r(t)$, we have that $\beta(v) = \alpha(v)$, i.e.,
		$\beta$ matches $\alpha$ on $t$'s read variables and;
		\item  $\beta\models r(t)$ and $\beta\models w(t)$,
		i.e., $\beta$ satisfies the read and write guards of $t$; and
		\item  for every $p \in \pre{t}$, it is the case that $M(p) \geq F(p,t)$.
	\end{compactitem}
	Given a transition $t$ enabled in state $(M,\alpha)$ under $\beta$,
	a state $(M',\alpha')$ is the result of \emph{firing $t$ in $(M,\alpha)$},
	written $(M, \alpha) \goto{t,\beta} (M', \alpha')$, iff:
	\begin{compactitem}	 	
		\item for every $p\in P$, we have $M'(p) = M(p) - F(p,t) + F(t,p)$; and
		\item for every $v\in \Var_w(t)$, we have $\alpha'(v)=\beta(v)$.
		\item for every $v\in V\setminus{\Var_w(t)}$, we have $\alpha'(v)=\alpha(v)$.
	\end{compactitem}
\end{definition}
We refer to the expression $(M, \alpha) \goto{t,\beta} (M', \alpha')$ as \emph{transition firing}.
State $(M',\alpha')$ is \emph{reachable} from $(M,\alpha)$,
if there exists a sequence of transition firings from
$(M,\alpha)$ to $(M',\alpha')$.

%A \emph{run} of $\net$ is a sequence
%$\eta=(M_0,\alpha_0)\goto{t_1, \beta_1}\cdots \goto{t_{\ell-1}, \beta_{\ell-1}} (M_\ell, \alpha_\ell)$
%of transition firings. We call $(M_0,\alpha_0)$ and $(M_\ell,\alpha_\ell)$, respectively, the
%\emph{initial} and \emph{final} state of $\eta$.

%%
%We denote with $\runs$ a set of all runs of $\net$.
%%
%Notice that, in general, $\net$ contains infinitely many states, due to
%the existence of infinitely many variable valuations.

%%By $\R(M,\alpha)$ we denote the set of all $\net$ states reachable from state $(M,\alpha)$.
%%Notice that $\R(M,\alpha)$ may contain infinitely many states due to
%%the existence of infinitely many variable valuations.

In this paper, we deal only with DPNs that are \emph{safe} (i.e., 1-bounded) and
\emph{well-formed (over their respective set of event signatures $\E$)}.
The former means that
%%, given a DPN $\net = (P, T, F, l, r, w)$,
for every state $(M',\alpha')$ reachable from a state $(M,\alpha)$,
if $M(p)\leq 1$ then $M'(p)\leq 1$. This is done for convenience (our approach seamlessly works for $k$-bounded nets).
%The latter means that, for every transition $t\in T$ and every
%event signature $\tup{n,A}\in\E$ s.t.~$l(t)=n$, it is the case that $\Var(r(t))\cup\Var(w(t))=A$
%(remember that no two signatures in $\E$ share the same name).
The latter means that transitions and event signatures are compatible, in the following sense:
\begin{inparaenum}[\itshape (i)]
\item for every (visible) transition $t\in T$ with $l(t)=n$ for
some event signature $\tup{n,A}\in\E$, we have that the write guard uses, as variables, precisely those matching with attributes in $A$, that is,
%$\Var(r(t))\cup
$\Var_w(t)=A$;
\item for every (silent) transition $t \in T$ with $l(t)=\tau$, net variables are left untouched, that is, $w(t)\equiv\top$.
\end{inparaenum}
The first requirement captures the intuition that the payload of an event is used to update the net variables, provided that the corresponding write guard is satisfied. The second requirement indicates that variables are only manipulated when a visible transition, triggered by an event, fires.

To define runs, we fix a \emph{DPN with initial state and final marking} (DPNIF) as a pair $\bar\net = (\net,(M_0,\alpha_0), M_f)$, where $\net$ is a DPN, $(M_0,\alpha_0)$
a state of $\net$ (called \emph{initial state}), and $M_f$ a marking of $\net$ (called \emph{final marking}). A \emph{run} of $\bar\net$ is a sequence of transition firings of $\net$ that starts from $(M_0,\alpha_0)$ and finally leads to a state $(M,\alpha)$ with $M = M_f$.

%$\eta=(M_0,\alpha_0)\goto{t_1, \beta_1}\cdots \goto{t_{\ell-1}, \beta_{\ell-1}} (M_\ell, \alpha_\ell)$
%
%
%
%A \emph{run} of $\net$ is a sequence
%of transition firings. We call $(M_0,\alpha_0)$ and $(M_\ell,\alpha_\ell)$, respectively, the
%\emph{initial} and \emph{final} state of $\eta$.

%We are interested in DPNs with fixed initial state and final marking.
%A
%We call $(M_0,\alpha_0)$ the \emph{initial state} of $\bar\net$, $M_f$ the \emph{final marking}
%of $\bar\net$, and every state $(M_f,\alpha)$ a \emph{final state} of $\bar\net$.
%A \emph{run of a DPINF $\bar\net$} is a run of $\net$ with initial state $(M_0,\alpha_0)$
%and final state $(M_f,\alpha)$.

We are now ready to define when a trace (in the sense of Def.~\ref{def:trace}) \emph{complies} with a DPNIF. This captures that the events contained in the trace can be turned into a corresponding run, possibly inserting $\tau$-transitions, while keeping the relative order of events and their correspondence to elements in the run. To do so, we need a preliminary notion. Given two sequences $\sigma_1$ and $\sigma_2$ such that $|\sigma_2| \geq |\sigma_1|$, an order-preserving injection $\iota$ from $\sigma_1$ to $\sigma_2$ is a total injective function from the elements of $\sigma_1$ to those of $\sigma_2$, such that for every two elements $e_1,e_2$ in $\sigma_1$ where $e_2$ comes later than $e_1$ in the $\sigma_1$, we have that $\iota(e_2)$ comes later than $\iota(e_1)$ in $\sigma_2$. This notion allow us to easily map traces into (possibly longer) runs of a DPNIF.
\begin{definition}\label{def:trace-dpn}
A trace $\sigma=e_1\cdots e_n$, \emph{complies} with a DPNIF $\bar\net$ with labeling function $l$ if there exist a run $\rho$ of $\bar\net$ and an order-preserving injection $\iota$ from $\sigma$ to $\rho$ such that:
\begin{compactitem}
\item every $e = \tup{n,\nu}$ in $\sigma$ is mapped by $\iota$ onto a corresponding transition firing in $\rho$, that is, given $\iota(e) = [t,\beta\rangle$, we have that $l(t) = n$ and $\beta$ corresponds to $\nu$ for the written variables $\Var_w(t)$;\footnote{Recall that $\beta$ involves both read and written variables. The read variables are used to guarantee that the fired transition is enabled, and it is on the written variables that $\nu$ and $\beta$ must agree.}
\item every element $[t,\beta\rangle$ in $\rho$ that does not correspond to any element from $\sigma$ via $\iota$ is so that $l(t) = \tau$.
\end{compactitem}
\end{definition}

\section{Monitoring Approach}
\label{sec:monitoring}

In this section we provide our main technical contribution: the construction of monitors for hybrid processes. In our context, a hybrid process $\H$ over a set $\E$ of event signatures is simply a set of process components, where each process component is either a \lmpdeclare constraint over $\E$, or a DPNIF over $\E$. Monitoring a trace against $\H$ basically amounts to running this trace concurrently over all the DPNIFs of $\H$, simultaneously checking whether all constraints in $\H$ are satisfied. When the trace is completed, it is additionally checked that the trace is indeed accepted by the DPNIFs. One important clarification is needed when characterizing the concurrent execution over multiple DPNIFs.
In fact, such components may come from different sources, not necessarily employing all the event signatures from $\E$. In this light, it would be counterintuitive to set that a DPNIF rejects an event because its signature is not at all used therein. We fix this by assuming that whenever such a situation happens, the DPNIF simply ignores the currently processed event.

Given this basis, the construction of monitors for such hybrid processes goes through multiple conceptual and algorithmic steps, detailed next.

%We proceed as follows. First, we show how we can reason on data conditions using a faithful, propositional abstraction based on intervals over the reals. Second, we introduce a finite-state automaton model tailored to our setting, namely natively supporting data conditions. Third, we combine these two techniques to show that LMP-\declare constraints and DPNs can be homogeneously characterised using automata of this type. Finally,

\subsection{Interval Abstraction}
\label{sec:abstraction}
The first challenge that one has to overcome is related to reasoning with data conditions, that is, checking whether a condition is satisfied by an assignment, and checking whether a condition is satisfiable (both operations will be instrumental when constructing automata).
The main issue is that, due to the presence of data, there are infinitely many distinct assignments from variables/attributes to values, in turn inducing infinitely many states to consider in the DPNs (even when the net is bounded). To tame this infinity, we build on the faithful abstraction techniques studied in \cite{LeoniFM18}, recasting them in our more complex setting. The idea is to avoid referring to single real values, and instead predicate over intervals, in turn showing that we only have a fixed number of intervals to consider, which in turn leads us to propositional reasoning.
%This is obtained by observing that data conditions have distinguishing power only over those  constants that are explicitly mentioned therein;
This is obtained by observing that data conditions can distinguish between only those constants that are explicitly mentioned therein;
hence, we simply fetch constants used in the process components (i.e., some atomic condition, guard or initial DPN assignment) to delimit the intervals to consider.

Technically, let $\C = \set{c_1,\ldots,c_m}$ be a finite set of values from $\reals$ assuming, without loss of generality,
that $c_i<c_{i+1}$, for $i \in \set{1,\ldots,m-1}$. We then partition $\reals$ into $\P_\C=\set{(-\infty,c_1),(c_m,\infty)}\cup\set{(c_i,c_i)\mid i=1,\ldots,m}\cup
\set{(c_i,c_{i+1})\mid i \in \set{1,\ldots,m-1}}$. Notice that $\P_\C$ is finite, with a size that is linear in $m$. This is crucial for our techniques: we can see $\P_C$ as a set of intervals over the reals or simply as a fixed set of propositions, depending on our needs.
Each interval in the partition is an \emph{equivalence region} for the satisfaction of the atomic conditions
$a\odot c$ in $\L_\E$, in the following sense: given two valuations $\alpha$ and $\alpha'$ defined over $a$,
such that $\alpha(a)$ and $\alpha'(a)$ are from the same region $R\in \P_\C$, then $\alpha\models v\odot c$ if and only if $\alpha'\models v\odot c$.

We exploit this as follows. We fix a finite set $V$ of variables (referring to attributes) and lift an assignment $\alpha:V \rightarrow \reals$ into a corresponding region assignment $\tilde\alpha:V\to \P_\C$ so that, for every $a \in V$, $\tilde\alpha(a)$ returns the unique interval to which $\alpha(a)$ belongs. Given the observation above, we can then use $\tilde\alpha$ to check whether a condition holds over $\alpha$ or not as follows: $\alpha(a)$ satisfies condition $a > c$ with $c \in \C$ if and only if $\tilde\alpha(a)$ returns a region $(c_1,c_2)$ with $c_1 > c$ (the same reasoning is similarly done for other comparison operators). This carries over more complex conditions used in \lmpdeclare and DPNs, as they simply consist of boolean combinations of atomic conditions. The key observation here is that doing this check amounts to propositional reasoning, and so does checking satisfiability of conditions: in fact, since both $V$ and $\P_\C$ are finite, there are only finitely many region assignments that can be defined from $V$ to $\P_\C$.

Given the process components of interest, we fix $V$ to the set $\A_\E$ of all the attributes in the event signature $\E$ of the system under study (this contains all variables used in its process components), and $\C$ to the set of all constants used in the initial states of the DPNs, or mentioned in some condition of a process component.
We then consistently apply the lifting strategy from assignments to region assignments, when it comes to traces and DPN states. In the remainder, we assume that $V$ and $\C$ are fixed as described above.

\subsection{Encoding into Guarded Finite-state Automata}
\label{sec:encoding}
As a unifying device to capture the execution semantics of process components, we introduce a symbolic automaton whose transitions are decorated with data conditions.

\begin{definition}%[Guarded Finite-state Automaton (\gfa)]
%%s
A \emph{guarded finite-state automaton} (\gfa) over set $\E$ of event signatures
is a tuple $\A=\tup{Q,q_0,\ra,F}$, where:
\begin{inparaenum}[\itshape (i)]
	\item $Q$ is a finite set of \emph{states};
	\item $q_0\in Q$ is the \emph{initial state};
	\item ${\lra}\subseteq Q\times\L_\E\times Q$ is the labeled \emph{transition function}; and
	\item $F\subseteq Q$ is the set of \emph{final states}.
\end{inparaenum}
For notational convenience, we write $q\ltr{\varphi}q'$ for $\tup{q,\varphi,q'}\in{\lra}$, and call $\varphi$ \emph{(transition) guard}.
\end{definition}
\gfa-runs of $\A$ consist of finite sequences of the form $q_0\ltr{\varphi_1}\cdots\ltr{\varphi_n}q_n$, where  $q_n \in F$. The set of runs accepted by $\A$ is denoted as $\L_\A$.
A trace $\sigma = e_1\cdots e_m$ over $\E$ is accepted by $\A$ if there exists a \gfa-run $q_0\ltr{\varphi_1}\cdots\ltr{\varphi_m}q_m$ such that for $i \in \set{1,\ldots,m}$, we have $e_i \models \varphi_i$.
In general, an event $e$ can satisfy the guards of many
transitions outgoing from a state $q$, as guards are not required to be
mutually exclusive. Thus, a trace may correspond to many \gfa-runs. In this sense,
\gfa{s} are, in general, nondeterministic.

It is key to observe that \gfa{s} can behave like standard finite-state automata. In fact, by setting $\C$ to a finite set of constants including all those mentioned in the automata guards, we can apply the interval abstraction from Section~\ref{sec:abstraction} to handle automata operations. In particular, in place of considering the infinitely many events over $\E$, we can work over the finitely many abstract events defined using region assignments over $\P_\C$. For example, we can check whether a trace $\sigma=\tup{n_1,\nu_1}\cdots\tup{n_m,\nu_m}$ is accepted by $\A$ by checking whether the abstract trace $\tup{n_1,\tilde\nu_1}\cdots\tup{n_m,\tilde\nu_m}$ does so.
Notice that, to construct this abstract trace, it suffices to represent each event $\tup{n,\nu}$ in $\sigma$ using equivalence regions from $\P_\C$,
such that every $\nu(a)=c$ is substituted either with region $[c,c]$, if $[c,c]\in \P_\C$, or with region $(c',c'')$ s.t. $c\in (c',c'')$. 
For ease of reference, we shall use %$e_{\P_\reals}$ to represent the modified event and 
$\sigma^{\P_\C}$ to represent the abstract trace.

Thanks to this, we can construct \gfa{s} using standard automata techniques (e.g., for \ltlf, as discussed below), and also directly apply standard algorithms, coupled with our interval abstraction, to minimize and determinize \gfa{s}.

\medskip
\noindent
\textbf{From LMP-\declare constraints to \gfa{s}.}
The translation of an \ltlf formula into a corresponding finite-state automaton \cite{DeVa13,DBLP:conf/bpm/GiacomoMGMM14} has been largely employed in the literature to build execution engines and monitors for \declare. In the case of \declare, atomic formulae are simply names of activities, and consequently automata come with transitions labeled by propositions that refer to such names. In the case of \lmpdeclare, atomic formulae are more complex conditions from $\L_\E$. Thanks to interval abstraction, this is however not an issue: we simply apply the standard finite-state automata construction for a \declare constraint \cite{DBLP:conf/bpm/GiacomoMGMM14}, with the only difference that transitions are labeled by those conditions from $\L_\E$ mentioned within the constraint. We only keep those transitions whose label is a satisfiable condition, which as discussed in Section~\ref{sec:abstraction} can be checked with propositional reasoning. A final, important observation is that the so-constructed \gfa is kept \emph{complete} (that is, \emph{untrimmed}), so that each of its states can process every event from $\E$.

%
% generating automata starting from \declare templates formali
%
%Let us now briefly discuss how \lmpdeclare constraints can be represented as GFAs.
%More concretely, given an \ltlf formula $\varphi$, we would like to construct a GFA $\A_\varphi$ that
%accepts exactly \emph{all traces} that satisfy $\varphi$.
%%%
%The algorithm for generating such automaton for an \ltlf formula without
%atomic conditions has been studied in literature~\cite{DeVa13} and there are
%also implemented automatic techniques that allow to construct it~\ref{AAAI17}.
%In our work, we do not deal  with the construction of GFAs for \lmpdeclare constraints
%and instead rely on the aforementioned algorithm with a proviso that,
%instead of labeling automaton transitions with propositions representing activity names only~\cite{DeVa13,AAAI17},
%we assume the algorithm working with conditions from $\L_\E$.
%The latter requires that for every variable in $\varphi$, one generates a formula similarly to Definition~\ref{def:ind-gfa-psi}
%that, contrary to analyzing a concrete DPN state, considers all equivalence regions derived from the constants from $\varphi$ for computing each $\phi_v$. Notice that each of such equivalence regions must satisfy $\varphi$.
%

\smallskip
\noindent
\textbf{From DPNIFs to \gfa{s}.} We show that a DPNIF $\bar\net = (\net,(M_0,\alpha_0),M_f) $  can be encoded into a corresponding \gfa that accepts all and only those traces that comply with $\bar\net$. For space reasons, we concentrate here on the most important aspects. 
%The interested reader can find details and algorithms in the companion report \cite{AAMP21}.

%We next define a finite-state automaton that accepts all and only the
%event traces that are compliant with a DPNIF $\bar\net$.
%
%%%When this is the case, the automaton is said to be
%%%nondeterminitisic (\gnfa), otherwise it is said to be deterministic (\gdfa).
%%%When not specified (\gfa), the automaton is assumed nondeterministic.
%%%In \gnfa's,
%
%Given a DPNIF $\bar\net=(\net,(M_0,\alpha_0),M_f)$, we can define a \gfa
%that accepts exactly all the event traces compliant with $\bar\net$.

%%

%\todo[inline]{$S$ has to be more specific: we need to consider only ``accepting'' runs that reach $M_f$ from $M_0$.}

The main issue is that the set $S$ of DPN states that are reachable from the initial marking
$(M_0,\alpha_0)$ is in general infinite even when the net is bounded (i.e., has boundedly many markings). This is due to the existence of infinitely many valuations for the net variables. To tame this infinity, we consider again the partition $\P_\C$ defined above; this induces a partition of $S$ into equivalence classes,
according to the intervals assigned to the variables of $\net$. Technically, given two assignments $\alpha,\alpha'$ we say that $\alpha$ is equivalent to
$\alpha'$, written $\alpha\sim\alpha'$, iff for every $v\in V$ there exists a region $R\in\P_\C$ s.t. $\alpha(v),\alpha'(v)\in R$.
Then, two states $(M,\alpha), (M',\alpha')\in S$ are said to be equivalent,
written $(M,\alpha)\sim (M',\alpha')$ iff $M=M'$ and $\alpha\sim\alpha'$.
Observe  that, by what discussed above, the assignments of two equivalent
states satisfy exactly the same net guards.
By $[S]_\sim$, we denote the quotient set of $S$ induced by the equivalence relation
$\sim$ over states defined above.

Based on Section~\ref{sec:abstraction} we directly get that $[S]_\sim$ is finite. We can then conveniently represent each equivalence
class of $[S]_\sim$ by $(M,\tilde\alpha)$, explicitly using the region assignment in place of the infinitely many corresponding value-based ones. This provides the basis for the following encoding.

\begin{definition}[\gfa induced by a DPNIF]\label{def:ind-gfa}
Given a DPNIF $\bar\net=(\net,(M_0,\alpha_0),M_f)$ with $\net = (P, T, F, l, r, w)$,
the \emph{\gfa induced by $\bar\net$} is
$\A_\net=\tup{Q,q_0,\ra,F}$, where:
\begin{compactenum}
	\item $Q= [S]_\sim$;
	\item $q_0=(M_0,\tilde\alpha_0)$, where
	$\tilde\alpha_0(v)=[\alpha_0(v),\alpha_0(v)]$, for all $v\in V$;
	\item \label{def:ind-gfa-state} ${\lra}\subseteq Q\times\L_\E\times Q$ is
	s.t.~$(M,\tilde\alpha)\ltr{a\land\psi}(M',\tilde\alpha')$ iff
	there exists a transition $t\in T$ and a partial valuation $\beta$
	s.t.~$(M,\alpha)\goto{t, \beta}(M',\alpha')$, with:
	\begin{compactenum}
		\item $\alpha(v)\in\tilde\alpha(v)$ and $\alpha'(v)\in\tilde\alpha'(v)$, for every $v\in V$;
		\item $a=l(t)$;
		\item \label{def:ind-gfa-psi} $\psi = \bigwedge_{v\in \Var(w(t))} \phi_v$
		such that:
		\begin{compactenum}[\itshape (i)]
			\item $\phi_v\equiv(v>c_i\land v<c_{i+1})$, if $\tilde\alpha'(v)=(c_i,c_{i+1})$;
			\item $\phi_v\equiv(v=c_i)$, if $\tilde\alpha'(v)=(c_i,c_i)$;
		\end{compactenum}
	\end{compactenum}	
	\item $F\subseteq Q$ is s.t.~$(M,\tilde\alpha)\in F$ iff $M=M_f$.
\end{compactenum}
\end{definition}
%This \gfa can be actually constructed using the algorithm provided in \cite{AAMP21}. 

Next we introduce an algorithm that, given a DPNIF $\bar\net= (\net,(M_0,\alpha_0),M_f)$, 
constructs the \gfa $\A_\net$ corresponding to it (that is, it represents all possible behaviors of $\bar\net$).
In the algorithm, we make use of the following functions:
\begin{itemize}[$\bullet$]
\item $enabled(M,\tilde\alpha)$ returns a set of transitions and region assignments 
$\set{(t,\tilde\beta)\mid  t\in T
\text{ and }
(M,\alpha)\goto{t,\beta}, 
\text{ where }\beta(v)\in\tilde\beta(v),\, 
\alpha(v)\in\tilde\alpha(v), \text{ for }v\in V}$.
%where $\tilde\beta$ is such that for any partial valuation $\beta$ and any variable valuation $\alpha$ as in Definition~\ref{def:ind-gfa},  it holds that $(M,\alpha)\goto{t,\beta}$. 
Notice that $\tilde\beta$ matches only the ``allowed'' regions. 
That is, for every $t$, we need to construct multiple $\tilde\beta$ 
that account for \emph{all possible combinations} of equivalence regions assigned to 
each variable in $w(t)$ and $r(t)$ such that 
$\tilde\beta\models w(t)$ and $\tilde\beta\models r(t)$.
\item $guard(t,\tilde\alpha)$ returns a formula $\psi$ as in Definition~\ref{def:ind-gfa-psi}.
\item $fire(M,t,\tilde\beta)$ returns a pair $(M,\tilde\alpha)$ as in Definition~\ref{def:ind-gfa-state}.
\end{itemize}
It is easy to see that all the aforementioned functions are computable. 
For $enabled$ there are always going to be finitely many combinations of 
regions from $\P_\reals$ satisfying the guards of $t$,
whereas formulas produced by $guard$ can be constructed using a version 
of the respective procedure from Definition~\ref{def:ind-gfa} 
that uses $\tilde\beta$ instead of $\tilde\alpha'$, and next states returned by $fire$ 
can be generated via the usual new state generation procedure from Definition~\ref{def:dpn-transition-firing}, 
proviso that it has to be invoked in the context of equivalence regions.

The actual algorithm is similar to the classical one used for the reachability graph construction of a Petri net (see, e.g.,~\cite{Murata89}).
It is important to notice that, in the proposed algorithm, silent transitions are treated as regular $\epsilon$-transitions (that is, we assume that $\L_\E$ as well as $\lra$ of the output automaton are suitably extended with $\tau$). 
We discuss later on how such transitions can be eliminated from resulting GFAs.

\begin{algorithm}
\caption{Compute GFA from DPN}
\begin{algorithmic} \label{alg:dpn-to-gfa}
\REQUIRE DPNIF $\bar\net=(\net,(M_0,\alpha_0),M_f)$ with $\net = (P, T, F, l, r, w)$
\ENSURE GFA $\tup{Q,q_0,\ra,F}$
	\STATE $Q:=(M_0,\tilde\alpha_0)$, where $\tilde\alpha_0(v)=[\alpha_0(v),\alpha_0(v)]$, for $v\in V$
	\STATE $\W:=\set{(M_0,\tilde\alpha_0)}$
	\STATE ${\lra}:=\emptyset$
	\WHILE{$\W\neq\emptyset$}
		\STATE select $(M,\tilde\alpha)$ from $\W$
		\STATE $\W:=\W\setminus (M,\tilde\alpha)$
		\FORALL{$(t,\tilde\beta)\in enabled(M,\tilde\alpha)$} 
			\STATE $(M',\tilde\alpha'):=fire(M,t,\tilde\beta)$
			\IF{$(M',\tilde\alpha')\not\in Q$}
				\STATE $Q:=Q\cup \set{(M',\tilde\alpha')}$
				\STATE $\W:=\W \cup \set{(M',\tilde\alpha')}$
			\ENDIF
			\STATE $\psi := l(t)$
			\IF{$w(t)\not\equiv\top$}
				\STATE $\psi := \psi\land guard(t,\tilde\alpha')$
			\ENDIF
			\STATE ${\lra}:={\lra}\cup\set{(M,\tilde\alpha)\ltr{\psi}(M',\tilde\alpha')}$
		\ENDFOR
	\ENDWHILE
\end{algorithmic}
\end{algorithm}

The following theorem outlines the main properties of the presented algorithm.
\begin{theorem}\label{thm:alg-soundness-termination}
Algorithm~\ref{alg:dpn-to-gfa} effectively computes 
a GFA $\A_\net$ induced by a DPNIF $\bar\net$, 
is sound and terminates.
\end{theorem}

However, Algorithm~\ref{alg:dpn-to-gfa} is not guaranteed to produce \gfa{s} that are complete. To ensure the completeness of the algorithm output, three additional modifications have to be performed. 
\begin{enumerate}
\item[\textbf{(M1)}] Given that the so-obtained \gfa $\A_\net$ does not properly handle silent, $\tau$-transitions (they are simply treated as normal ones by the algorithm), we need to compile them away from $\A_\net$, which is done by recasting the standard $\epsilon$-move removal procedure for finite-state automata based on $\epsilon$-closures (see, e.g., \cite{automata-theory}) into our setting. 
This procedure allows to collapse (sequences of) states corresponding to $\tau$-transitions into those that are used for representing only behaviors of all possible traces complying to $\E$. \footnote{Since we assume to be working with well-formed nets only, $\tau$-transitions do not come with write guards, and therefore do not alter the net variables. }
To collapse the aforementioned sequences, each state $q\in Q$ s.t.
there are $k$ runs $q\ltr{\tau}\cdots \ltr{\tau}q_i'$, where every transition is labeled only with $\tau$, $q_i'\ltr{\varphi}q_i''$, $\varphi\neq \tau$ and $i=1,\ldots,k$, 
has to be replaced with the set of all such $q_i'$ states by creating additional transitions to predecessors of $q$ (if any) as well as $q_i''$. 
Notice also that, while the $\epsilon$-transition removal procedure produces deterministic automata, its counterpart working with GFAs may produce an automaton that is non-deterministic.  
\item[\textbf{(M2)}] The output \gfa has to be made ``tolerant'' to events whose signature is not at all used in the DPNIF, formalising the intuition described at the beginning of Section~\ref{sec:monitoring}. This is done by introducing additional loops in $\ra_i$ from Definition~\ref{def:ind-gfa} as follows: for every $q\in Q_i$ we insert a looping transition $q\ltr{\psi}q$, where $\psi=\bigwedge_{a\in(\N_\E\setminus \bigcup_{t\in T} l(t))}a$. By doing so, we allow the \gfa to skip irrelevant events that could never be processed by the net. 
\item[\textbf{(M3)}] The resulting \gfa has to be extended with two types of extra transitions. 
	\begin{itemize}
		\item The first one tackles invalid (or incorrect) net executions where a partial run cannot be completed into a proper run due to a data-related deadlock. This can occur for two reasons, respectively related to the violation of read versus write guards of all transitions that, from the control-flow perspective, could actually fire.
To deal with the read-related issue we proceed as follows: for every state $(M,\tilde\alpha)\in Q_i$ 
and every transition $t\in T_i$, if $\tilde\alpha\not\models r(t)$ and $M(p)\geq F_i(p,t)$ for every $p\in P_i$,
%then $\ra_i:=\ra_i\cup (M,\tilde\alpha)\ltr{\psi}(M,\tilde\alpha)$,
then add $(M,\tilde\alpha)\ltr{\psi}(M,\tilde\alpha)$ to $\ra_i$,
where $\psi$ is as in Definition~\ref{def:ind-gfa} (see \ref{def:ind-gfa-state}). This is only done when $\psi$ is actually satisfiable.
	\item The second one addresses write-related issues arising when the event to be processed carries values that violate all the write guards of candidate transitions. We handle this as follows: for every $(M,\tilde\alpha)\in Q_i$ 
and every $t\in T_i$ s.t.~$w(t)\not\equiv \top$, 
add $ (M,\tilde\alpha)\ltr{a\land\psi}(M,\tilde\alpha)$ to $\ra_i$,
where, for $t\in T_i$ and every $p\in\pre{t}$ s.t. $M(p)\geq F(p,t)$, $a=l(t)$
and $\psi$ is constructed as in Definition~\ref{def:ind-gfa} (bullet \ref{def:ind-gfa-state},  with the only 
difference that all $\varphi_v$ are computed for a combination of 
equivalence regions from $\P_\C$, each of which is composed into (partial)  
variable region valuation $\tilde\beta: V \to\P_\C$ 
s.t. $\tilde\beta\not\models v\odot c$, for every atomic condition $v\odot c$
in $w(t)$. This is only done if $\psi$ is actually satisfiable. 
This step can be also optimised by putting all such $\psi$ in one DNF formula,
which in turn reduces the number of transitions in the \gfa.
%Notice that the so-added transitions ensure that  \emph{all possible combinations} of intervals that violate $w(t)$ are considered. 
	\end{itemize} 
\end{enumerate}

\begin{proposition}
Let $\A_\net$ be a GFA  induced by a DPNIF $\bar\net$.
Application of modifications \textbf{(M1)}, \textbf{(M2)} and \textbf{(M3)} to $\A_\net$ produces a new GFA $\A_\net'$ that is complete.
\end{proposition}

%\begin{IEEEproof}
%\end{IEEEproof}

Now, whenever we get a complete GFA for a DPNIF, we can use the former to check whether a log trace is compliant with the net (as by Definition~\ref{def:trace-dpn}). 

\begin{theorem}\label{thm:abstraction}
	A trace $\sigma=e_1\cdots e_n$ is compliant with a 
	DPNIF $\bar\net= (\net,(M_0,\alpha_0),M_f)$ iff 
%	$\sigma$ is accepted by the 
	abstract trace $\sigma^{\P_\C}$ is accepted by the 
	\gfa $\A_\net$ induced by $\bar\net$.
\end{theorem}

%As detailed in \cite{AAMP21}, three additional steps are performed to obtain a final \gfa that has the property of being complete. The first step compiles away $\tau$-transitions, recasting standard techniques for removing $\epsilon$-moves in automata. The second step extends the \gfa with self-loops to skip/ignore all those events whose signature is outside the one used in the DPNIF (formalizing the intuition described at the beginning of Section~\ref{sec:monitoring}). The last step is about completing the \gfa with further transitions dealing with events that are not accepted by the DPNIF. These all lead to a non-accepting, trap state.

%These are transitions from ``data-deadlocking'' non-accepting states where none of the read guards evaluates to true, or transitions corresponding to events that violate the write guards of enabled transitions.

\subsection{Combining \gfa{s}}
Given a hybrid process $\H = \set{h_1,\ldots,h_n}$ with $n$ components, we now know how to compute a \gfa for each of its components. Let $\A_i$ be the \gfa obtained as described in Section~\ref{sec:encoding}, depending on whether $h_i$ is a \lmpdeclare constraint or DPNIF, and in addition minimized and determinized. This means that, being $\A_i$ complete, it will have a single trap state capturing all those traces that permanently violate the process component.

%
%
%In our monitoring setting, we assume that there can be multiple DPNs as well as \lmpdeclare
%constraints characterizing the same system/scenario. \todo{say more about the general scenario}
%To this end, we consider a set $\Phi=\set{\tup{\A_1,c_1}\ldots \tup{\A_n,c_n}}$, where each
%$\A_i=\tup{Q_i,q_{i0},\ra_i,F_i}$ is a \gfa associated with an input constraint or a DPN,
%and $c_i$ the corresponding cost incurred for violating it.
%We also require that each $\A_i$ is \emph{complete}, that is, it can process all possible events coming from an event log in which events have signatures from $\E$, deterministic  and \emph{minimal}, that is, it has the minimal number of states.
%The last two can be achieved by employing one of the already well-known canonization algorithms in which minimality can be achieved by producing a new GFA with one of the existing minimaztion algorithms for finite automata (see, e.g., \cite{hopcroft}).

To perform monitoring, we follow the approach of colored automata \cite{MMWV11,DBLP:conf/bpm/GiacomoMGMM14} and
 label each automaton state with one among fourth truth values, respectively indicating whether, in such state, the corresponding process component is \emph{temporarily satisfied} (TS), \emph{temporarily violated} (TV), \emph{permanently satisfied} (PS), or \emph{permanently violated} (PV). As for constraints, these are interpreted exactly like in \cite{MMWV11,DBLP:conf/bpm/GiacomoMGMM14}. As for DPNIF, TS means that the current trace is accepted by the DPNIF, but can be extended into a trace that is not, while TV means that the current trace is a good prefix of a trace that will be accepted by the DPNIF (PV and PS are defined dually).

Considering the way our \gfa{s} are obtained, this labeling is done as follows:
\begin{inparaenum}[\itshape (i)]
	\item $v_i(q_i)=PS$ iff $q_i\in F_i$ and all transitions
		outgoing from $q$ are self-loops;
	\item $v_i(q_i)=TS$ iff $q_i\in F_i$ and there is some transition outgoing from
		$q$ that is not a self-loop;
	\item $v_i(q_i)=PV$ iff $q_i\notin F_i$ and all transitions
		outgoing from $q$ are self-loops;
	\item $v_i(q_i)=TV$ iff $q_i\notin F_i$ and there is some transition outgoing from
		$q_i$ that is not a self-loop.
\end{inparaenum}
%%
%Specifically, the assignments $TS,TV,PS,PV$ to $q_i\in Q_i$ capture the
%situation where $q_i$ is a state of $\A_i$ from which constraint $\varphi_i$ (resp., net $\net_i$)
%is, respectively, \emph{temporarily satisfied},
%\emph{temporarily violated}, \emph{permanently satisfied} and
%\emph{permanently violated}.
%Notice that, given that automata are assumed to be complete,
%$v_i$ labels the states in such a way that the satisfiability or violation
%value will be correctly represented for any trace.

%Let us now look into additional requirements that have to be respected by every single
%automaton from $\Phi$.
%First of all, we require that every such automaton is constructed
%by taking into consideration larger equivalence regions that accommodate constants from
%all DPNs and \lmpdeclare constraints whose induced automata are in $\Phi$. This does not
%alternate the construction procedures discussed in the previous section and
%is needed for the correct representation of all possible values that can be handled
%by every automaton from $\Phi$ in the final product automaton.

The so-obtained labeled \gfa{s} are local monitors for the single process components of $\H$. To monitor $\H$ as a whole and do early detection of violations arising from conflicting components, we need to complement such automata with a global \gfa $\A$, capturing the interplay of components. We do so by defining $\A$ as a suitable \emph{product automaton}, obtained as
a cross-product of the local \gfa{s}, suitably annotated to retain some relevant
information.

Technically, $\A=\tup{Q,q_0,\ra,F}$, where:
\begin{inparaenum}[\itshape (i)]
	\item $Q=Q_1\times\cdots\times Q_n$;
	\item $q_0=\tup{q_{10},\ldots,q_{n0}}$;
	\item $\lra$ is s.t.~$\tup{q_1,\ldots,q_n}\ltr{\varphi}\tup{q'_1,\ldots,q'_n}$ iff
		$\varphi=\varphi_1\land\cdots\land\varphi_n$, with
		$q_i\ltr{\varphi_i}_iq_i'$ and $\varphi$ is satisfiable by exactly one
		event;
	\item $F=F_1\times\cdots\times F_n$.
\end{inparaenum}
%\begin{itemize}
%	\item $Q=Q_1\times\cdots\times Q_n$;
%	\item $q_0=\tup{q_{10},\ldots,q_{n0}}$;
%	\item $\lra$ is s.t.~$\tup{q_1,\ldots,q_n}\ltr{\varphi}\tup{q'_1,\ldots,q'_n}$ iff
%		$\varphi=\varphi_1\land\cdots\land\varphi_n$, with
%		$q_i\ltr{\varphi_i}_iq_i'$ and $\varphi$ is satisfiable by exactly one
%		event;
%	\item $F=F_1\times\cdots\times F_n$.
%\end{itemize}
%Intuitively, $\A$ is the synchronous cross-product of all $\A_i$'s , i.e., an automaton that
%simulates the concurrent execution of the $\A_i$'s on the same log trace.
Observe that the
definition requires checking whether guards of $\A$ are satisfiable by some event
(otherwise the labeled transition could not be triggered by any event and should be omitted).
%While this is in general an NP-hard problem, the involved formulas are
%typically short in practice, thus greatly simplifying the check.

%We next show how to use $\A$ to do monitoring, identify conflicting constraints and find
%the events that allow for cost minimization.

The $v_i$ labeling functions from above induce a labeling
on the states $q=\tup{q_1,\ldots,q_n}\in Q$ of the product
automaton, which tells us whether all constraints and nets nets from $\Phi$ are overall
temporarily/permanently violated/satisfied. Specifically, we define the
labeling $v:Q\mapsto\set{TS,TV,PS,PV}$ s.t.:
\begin{inparaenum}[\itshape (i)]
 \item $v(q)=PV$ iff $v_i(q_i)=PV$, for some $i\in{1,\ldots,n}$;
 \item $v(q)=PS$ iff $v_i(q_i)=PS$, for all $i\in{1,\ldots,n}$;
 \item $v(q)=TS$ iff $v_i(q_i)=TS$, for all $i\in{1,\ldots,n}$;
 \item $v(q)=TV$, otherwise.
\end{inparaenum}

\subsection{Best Event Identification}	
It is crucial to notice that, differently from local \gfa{s}, the global \gfa $\A$ is \emph{not} minimized. This allows the monitor to distinguish among states of permanent violations arising from different combinations of permanently violated components, in turn allowing for fine-grained feedback on what are the ``best'' events that could be processed next. To substantiate this, we pair a hybrid process $\H$ with a \emph{violation cost function} that, for each of its components, returns a natural number indicating the cost incurred for violating that component.

$\A$ is augmented as follows. Each state $q\in Q$ of $\A$ is associated with two cost indicators:
\begin{compactitem}
	\item a value $\ccur(q)$ containing the sum of the costs associated with the
	constraints violated in $q$;
	\item a value $\cbest(q)$ containing the best value $\ccur(q')$,
	for some state $q'$ (possibly $q$ itself) reachable from $q$.
%%	\item a tuple $v(q)=\tup{v_1,\ldots,v_n}$, with $v_i\in\set{PS,PV,TS,TV}$
%%		expressing whether the automaton $\A_i$ is in a state $q_i$
%%		that yields permanent satisfaction ($PS$) (resp.~violation, $PV$)
%%		or temporary satisfaction ($TS$) (resp.~$TV$) of the respective
%%		constraint $\varphi_i$;
%%		
%%	\item a value $v\in\set{PS,PV,TS,TV}$ expressing whether the (compound)
%%		state $q$ yields a permanent or temporal satisfaction (resp.~violation)
%%		of all (resp.~some) constraints;
		
\end{compactitem}
The functions $\ccur$ and $\cbest$ can be easily computed through the fixpoint
computation described below, where $c_i$ is the cost associated with
the violation of constraint $\varphi_i$.
\begin{compactenum}
	\item For every state $q=\tup{q_1,\ldots,q_n}\in Q$, let
		\[\cbest^0(q)=\ccur(q)=\sum_{i=1,\ldots,n}\cfunc(q_i),\]
		where $\cfunc(q_i)=0$ if $q_i\in F_i$, $c_i$ otherwise.
%		\[
%			\cfunc(q_i)=
%			\begin{cases}
%				0 & \text{if }q_i\in F_i\\
%				c_i & \text{otherwise}
%			\end{cases}
%		\]

	\item\label{alg:2} Repeat the following until $\cbest^{i+1}(q)=\cbest^i(q)$,
	for all $q\in Q$:
		for every state $q\in Q$,
				\[\cbest^{i+1}(q):=
					\min\set{{\cbest^i(q')}\mid{q\ra q'}}\cup\set{\ccur(q)};\]
	\item return $\cbest$.
\end{compactenum}
It is immediate to see that a fixpoint is eventually reached in finite time,
thus the algorithm terminates.
To this end, observe that,
for all $q\in Q$, $\cbest(q)$ is integer and non-negative. Moreover,
at each iteration, if $\cbest(q)$ changes, it can only decrease. Thus, after
a finite number of steps, the minimum $\cbest(q)$ for each
state $q\in Q$ must necessarily be achieved, which corresponds to the
termination condition.

We can also see that the algorithm is correct, i.e., that if $\cbest(q)=v$
then, from $q$:
\begin{inparaenum}[\itshape (i)]
	\item there exists a path to some state $q'$ s.t.~$\ccur(q')=v$, and
	\item there exists no path to some state $q''$ s.t.~$\ccur(q'')<v$.
\end{inparaenum}
 These come as a consequence of step~\ref{alg:2} of the algorithm. By this,
indeed, we have that, after the $i$-th iteration $\cbest(q)$ contains the
value of the state with the minimum cost achievable from $q$ through
a path containing at most $i$ transitions.
When the fixpoint is reached, that means that even considering longer paths
will not improve the value, that is, it is minimal.

Using $\A$, $\ccur$ and $\cbest$, we can find the next ``best events'', i.e.,
those events that allow for satisfying the combination of constraints that
guarantees the minimum cost. Technically, let $\sigma=e_1\cdots e_\ell\in\EV^*$
be the input trace and consider the set
$\Gamma=\set{\rho_1,\ldots,\rho_n}$ of the runs of $\A$ on $\sigma$.
Let $q_{i\ell}$ be the last state of each $\rho_i$ and
let \[\hat q=\argmin_{q\in \set{q_{1\ell},\ldots,q_{n\ell}}}\set{\cbest(q)}.\]

If, for some $i$, $\hat q=q_{i\ell}$, then $\hat q$ is the best achievable
state and no event can further improve the cost.
Otherwise, take a successor $q'$ of $\hat q$ s.t.~$\cbest(q')=\cbest(\hat q)$.
Notice that by the definition of $\cbest$, one such $q'$
necessarily exists, otherwise $\hat q$ would have a different cost
$\cbest(\hat q)$.

The best events are then the events $e^*$ s.t.:
\[e^*\models \varphi\text{, for } \varphi \text{ s.t.~} q\ltr{\varphi}q'\]

Notice that, to process the trace in the global \gfa $\A$, and to detect the best events, we again move back and forth from traces/events and their abstract representation based on intervals, as discussed in Section~\ref{sec:abstraction}. In particular, notice that there may be infinitely many different best events, obtained by different groundings of the attributes within the same intervals.

\section{Conclusion}
\label{sec:conclusion}

The ability to monitor the interplay of different process models is useful in domains where process instances tend to have high variability. An example of this is the medical domain where standard treatment procedures are described as clinical guidelines and multiple guidelines need to be often executed simultaneously, therefore giving rise to interplay and possible conflicts. Furthermore, because a clinical guideline cannot account for all possible preconditions that a patient may have, it is also necessary to employ declarative knowledge (allergies, prior conditions etc.) which further complicates the process execution.

This paper proposes a monitoring approach that can take into consideration the interplay of multiple process specifications (both procedural and declarative). Additionally, the approach includes a recommendation component which helps either to avoid violations or (if avoiding the violations is not possible) to minimize the total cost of the violations. The proposed approach is limited in that it can only provide recommendations on the immediate next events, however in the future we plan to extend the approach to provide recommendations as full continuations of the trace and to explore different possible execution semantics for concurrent execution of multiple process models.

\bigskip
\noindent\textbf{Acknowledgements.}
The work of A.\ Alman was supported by the \mbox{Estonian} Research Council (project PRG1226) and ERDF via the IT Academy Program.

\bibliographystyle{abbrv}
\bibliography{bibliography}

\end{document}